\documentclass[letterpaper]{article} 
\usepackage[draft]{aaai2026}  
\usepackage{times}  
\usepackage{helvet}  
\usepackage{courier}  
\usepackage[hyphens]{url}  
\usepackage{graphicx} 
\usepackage{natbib}  
\usepackage{caption} 
\usepackage{algorithm}
\usepackage{algorithmic}

\usepackage{newfloat}
\usepackage{listings}
\DeclareCaptionStyle{ruled}{labelfont=normalfont,labelsep=colon,strut=off} 
\floatstyle{ruled}
\newfloat{listing}{tb}{lst}{}
\floatname{listing}{Listing}
\usepackage{verbatim}
\usepackage{amsmath}
\usepackage{amsthm}
\usepackage{tabularray}
\newcommand{\conf}{\mathit{Conf}}

\usepackage{xcolor}
\usepackage{soul}

\newcommand{\cruxwogc}{\text{CRUX}_{\neg\text{GC}}}

\title{A Context-Aware Dual-Metric Framework for Confidence Estimation in Large Language Models}

\author {
    Mingruo Yuan\textsuperscript{\rm 1},
    Shuyi Zhang\textsuperscript{\rm 2}*,
    Ben Kao\textsuperscript{\rm 1}
}
\affiliations {
    \textsuperscript{\rm 1}The University of Hong Kong\\
    \textsuperscript{\rm 2}Reliability Lab, Huawei Technologies Ltd.\\
    mingruo.yuan@connect.hku.hk, zhangshuyi13@huawei.com, kao@cs.hku.hk
}

\usepackage{bibentry}

\begin{document}

\maketitle
\begin{abstract}
Accurate confidence estimation is essential for trustworthy large language models (LLMs) systems, as it empowers the user to determine when to trust outputs and enables reliable deployment in safety-critical applications. Current confidence estimation methods for LLMs neglect the relevance between responses and contextual information, a crucial factor in output quality evaluation, particularly in scenarios where background knowledge is provided. To bridge this gap, we propose \textbf{CRUX} (Context-aware entropy Reduction and Unified consistency eXamination), the first framework that integrates context faithfulness and consistency for confidence estimation via two novel metrics. First, \textit{contextual entropy reduction} represents data uncertainty with the information gain through contrastive sampling with and without context.  Second, \textit{unified consistency examination} captures potential model uncertainty through the global consistency of the generated answers with and without context. Experiments across three benchmark datasets (CoQA, SQuAD, QuAC) and two domain-specific datasets (BioASQ, EduQG) demonstrate CRUX's effectiveness, achieving the highest AUROC than existing baselines.

\end{abstract}
\section{Introduction}
Large language models (LLMs) are widely deployed in real-world scenarios that commonly involve contextual question answering (CQA) tasks \cite{cqa_1, cqa_2}. Crucially, in these tasks, generating accurate responses fundamentally hinges on faithful interpretations of the provided context. However, the probabilistic nature of LLM inevitably introduces hallucinations \cite{NEURIPS2024_3c1e1fdf, bang2025hallulensllmhallucinationbenchmark} or errors, even when task-specific information is explicitly provided \cite{23hallucination_e, 25hallucination_e}. For example, in the legal domain, an LLM might disregard user-provided details and generate erroneous advice that contradicts the given terms, potentially leading to significant consequences.

To address these reliability challenges, researchers have developed various confidence estimation approaches \cite{liu2024uncertaintyestimationquantificationllms,ling2024uncertaintyquantificationincontextlearning, he2025surveyuncertaintyquantificationmethods}. Current methods primarily rely on consistency-based methods \cite{su, ecc, luq} (e.g., measuring answer variation across multiple samplings) or self-evaluation \cite{lin2022teachingmodelsexpressuncertainty, xiong2024llmsexpressuncertaintyempirical, heo2025llmsestimateuncertaintyinstructionfollowing} (e.g., prompting LLMs to assess their own certainty). While these approaches offer partial insights, self-evaluations are inherently untrustworthy due to the tendency of LLMs toward overconfidence \cite{24overconfidence, 25overconfidence}, and consistency alone allows models to generate consistently ungrounded or incorrect answers by relying solely on parametric knowledge that ignores or contradicts the evidence in the source input \cite{shi-etal-2024-trusting}. 
\begin{figure*}
    \centering
    \includegraphics[width=0.78\linewidth, height = 4.5cm]{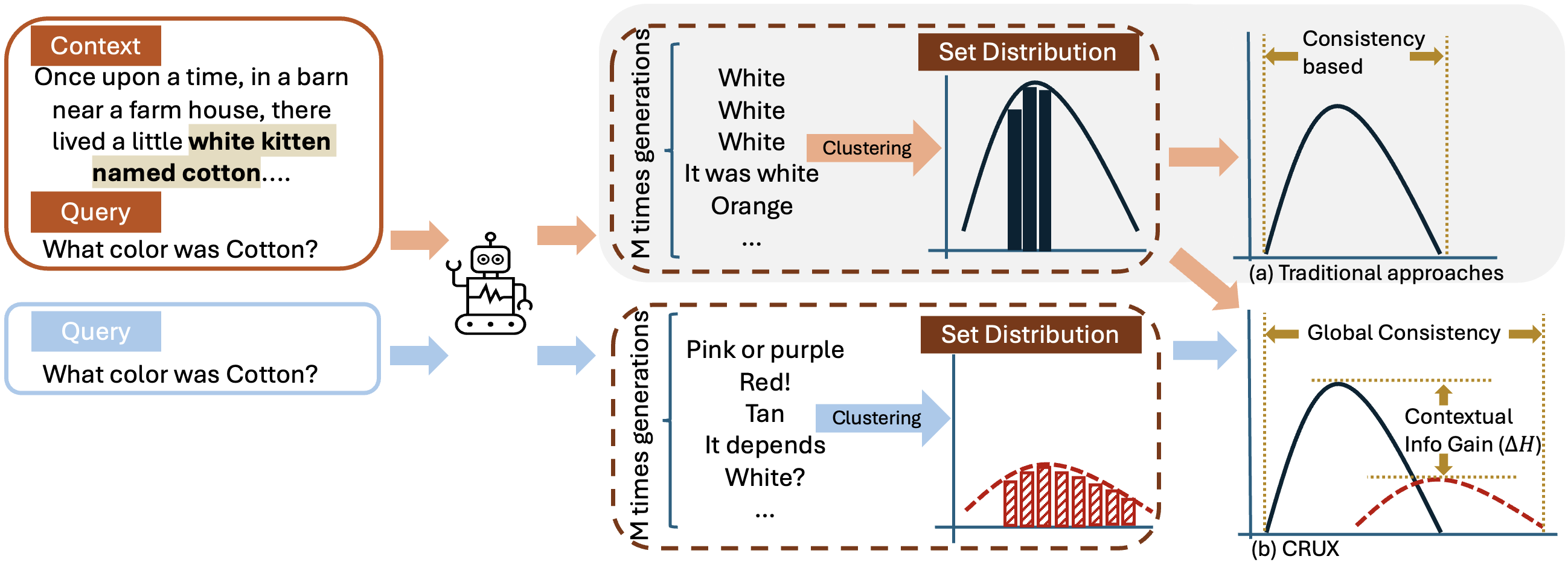}
    \caption{Illustration of Traditional Consistency vs. CRUX Methodology: \textbf{(a)} Conventional approaches focus exclusively on response consistency. \textbf{(b)} The CRUX framework enhances evaluation by combining contextual faithfulness (assessed via contextual information gain) with global consistency.}
    \label{fig:method}
\end{figure*}

This reveals a fundamental limitation in the field: prevailing confidence estimation techniques assess confidence primarily based on the stability or self-consistency of the model's responses. Crucially, this assessment is largely decoupled from the specific source input (e.g., the context in CQA tasks) that should ground the generation. Consequently, these methods fail to capture whether the outputs truly align with and are justified by the information contained within the provided source input, which is the very foundation of trustworthy generation systems. In CQA scenarios specifically, this means they cannot evaluate whether the model's answer faithfully reflects the given context.

To address this gap, we propose CRUX, a dual-metric framework that quantifies predictive confidence by combining contextual faithfulness and unified consistency, two dimensions essential for robust uncertainty estimation. Specifically, we first introduce a contrastive sampling strategy to measure how effectively an LLM utilizes contextual information. A significant entropy reduction when context is removed indicates that the context meaningfully constrains outputs and mitigates input knowledge gaps. In this way, entropy reduction can also be regarded as a measure for data uncertainty.

In contrast, a negligible entropy reduction suggests the model relied solely on its inherent knowledge or biases, overlooking the provided context. This renders the context irrelevant to the answer, which can happen in two distinct ways:
(1) low model uncertainty: The model already possesses sufficient internal knowledge to answer correctly inherently, independent of the context, or (2) high model uncertainty: The model lacks sufficient parametric knowledge to answer correctly even with the provided context, indicating that it failed to effectively utilize or comprehend the context.

To disentangle these two cases, we introduce unified consistency (also denoted as global consistency), which quantifies output stability under both contextual and context-free scenarios. High consistency implies the model robustly converges to correct answers, suggesting high confidence and low model uncertainty, while low consistency exposes high model uncertainty. 

In short, our approach explicitly evaluates both answer consistency and contextual dependency through a dual-path verification where responses must simultaneously agree with each other and remain anchored to the source context when the context itself is informative and relevant.

Our contributions are threefold:
\begin{itemize}
\item[*] We propose the \textbf{first} confidence estimation framework for CQA that jointly integrates \textbf{contextual faithfulness} and \textbf{unified consistency}. This multi-dimensional approach fundamentally advances the robustness of confidence assessment in context-dependent settings.
\item[*] We design a novel technique to quantify contextual faithfulness using contrastive sampling. This measures entropy between outputs generated with and without context. The resulting entropy reduction directly measures the model's reliance on and faithfulness to the context, serving as a core confidence indicator.

\item[*] We introduce unified consistency as a complementary metric. This allows us to decouple and identify scenarios dominated by model uncertainty. Crucially, this decomposition enables the diagnosis of targeted errors to distinguish failures due to model limitations where model improvement is required.

\end{itemize}



\section{Background}
Quantifying confidence in Natural Language Generation (NLG) systems is essential for reliable deployment. Unlike classification tasks \cite{NEURIPS2019_8558cb40} with finite outputs, NLG confidence estimation must account for the combinatorial nature of text generation. Predictive uncertainty quantification \cite{NEURIPS2018_3ea2db50,aichberger2024rethinkinguncertaintyestimationnatural} is a cornerstone of trustworthy language modeling. In NLG systems, this translates to measuring the reliability of generated sequences, which can be expressed using entropy: 
\begin{equation}
    -H(S|x) = \sum_{s \in \mathcal{S}} p(s|x) \log p(s|x),
	\label{eq:confidence}
\end{equation}
where $x$ denotes the input, $s$ is a possible output sequence, and $\mathcal{S}$ the space of all valid sequences. This formulation quantifies output distribution concentration. High confidence corresponds to peaked distributions (low entropy), while low confidence indicates dispersed probabilities (high entropy).

Two fundamental uncertainty types underlie this total uncertainty measure \cite{osband2023epistemicneuralnetworks, johnson2024expertsdontcheatlearning, NEURIPS2024_6aebba00}:
\begin{itemize}
	\item \textbf{Model uncertainty}: Arises from parameter estimation errors and knowledge gaps of the model. Reducible through improved architectures or additional training data.
	\item \textbf{Data uncertainty}: Stems from inherent ambiguities in inputs (e.g., multiple valid interpretations of ``light'' as weight or brightness). Irreducible without fundamentally different information.
\end{itemize}
A shortcoming of using Equation~\ref{eq:confidence} to measure uncertainty is that it is based solely on model output ($s$), which makes it difficult to isolate the two constituting uncertainty elements (Model and Data). In this paper we propose a new confidence metric $\conf(x)$ that takes model input into account, assesses both model uncertainty and data uncertainty, and integrates them into a single measure that reflects the interplay between the two types of uncertainty.
It thus serves as a diagnostic tool that quantifies data uncertainty and disentangles model uncertainty through its dual components. 

\section{Method}
\label{crux}
\subsection{Context-aware Entropy Reduction and Unified Consistency Examination}
Our method quantifies confidence in CQA tasks by integrating contextual faithfulness and unified consistency. It operates in three stages: (1) Compute contextual information gain via contrastive sampling and measuring entropy reduction. (2) Measure unified consistency under contextual and context-free scenarios. (3) Dynamic weighting to fuse metrics into a final confidence score. 
\begin{algorithm}[tb]
	\caption{CRUX}
	\small
	\label{alg:crux}
	\textbf{Input}: Query $q$, Context $c$, Sample size $n$\\
	\textbf{Output}: Confidence score $\conf \in [0,1]$
	\begin{algorithmic}[1]
		\STATE $//$ \textbf{Stage 1: Contextual Information Gain}
		\STATE \quad \textit{Contrastive Generation:}
		\STATE \quad $A^{(c,q)} \gets \{a_i \sim P(a \mid c, q)\}_{i=1}^n$ 
		\STATE \quad $A^{(q)} \gets \{a_i \sim P(a \mid q)\}_{i=1}^n$ 
		\STATE \quad Apply bidirectional entailment clustering:
		\STATE \quad \quad $K^{(c,q)} \gets$ partition $A^{(c,q)}$ into $\alpha$ semantic clusters
		\STATE \quad \quad $K^{(q)} \gets$ partition $A^{(q)}$ into $\beta$ semantic clusters
		
		\STATE \quad \textit{Entropy Reduction:}
		\STATE \quad $H(K^{(c,q)}) \gets -\sum_{k \in K^{(c,q)}} P(k|c,q) \log P(k|c,q)$
		\STATE \quad $H(K^{(q)}) \gets -\sum_{k \in K^{(q)}} P(k|q) \log P(k|q)$

		\STATE \quad $\Delta H \gets H(K^{(q)}) - H(K^{(c,q)})$
		
		\STATE $//$ \textbf{Stage 2: Unified Consistency Measurement}
		\STATE \quad $A^{\text{global}} \gets A^{(c,q)} \cup A^{(q)}$ 
		\STATE \quad Construct graph $G$ with nodes $A^{\text{global}}$ and edge weights as semantic similarities
		\STATE \quad Compute unified consistency:
		\STATE \quad \quad $GC_{\text{pairwise}} \gets \frac{1}{n(1-2n)} \sum_{1 \leq i<j \leq 2n} d(a_i^{\text{global}}, a_j^{\text{global}})$ 
		\STATE \quad \textbf{or} 
		\STATE \quad \quad $GC_{\text{center}} \gets -\frac{1}{2n} \sum_{i=1}^{2n} d(A_{\text{center}}, a_i^{\text{global}})$
		
		\STATE $//$ \textbf{Stage 3: Neural Weighting}
		\STATE \quad $\mathbf{v} \gets [\Delta H; GC]$  
		\STATE \quad $\mathbf{h} \gets \operatorname{ReLU}(W_1 \mathbf{v} + b_1)$
		\STATE \quad $\mathbf{o} \gets W_2 \mathbf{h} + b_2$
		\STATE \quad $\conf \gets \sigma(\mathbf{o})$ 
		
		\STATE \textbf{return} $\conf$
	\end{algorithmic}
\end{algorithm}
Algorithm~\ref{alg:crux} outlines the procedure.

The objective of the first stage is to quantify how much an LLM relies on provided context. We represent confidence by contextual information gain. Given a query $q$ and its associated context $c$, our CRUX framework proceeds as follows:

\paragraph{Contrastive Generation}
We first sample $n$ answers conditioned on $q$ and $c$: ${A}^{(c,q)}=\left\{a_1^{(c,q)},a_2^{(c,q)},\ldots,a_n^{(c,q)}\right\}$, where $a_i^{(c,q)}\sim P(a\mid c,q)$. Given complete information, we expect the model to generate answers that are both semantically consistent with each other and grounded in the provided context.
Then, we sample the same number of answers conditioned only on $q$: ${A}^{(q)}=\left\{a_1^{(q)},a_2^{(q)},\ldots,a_n^{(q)}\right\}$, where $a_i^{(q)}\sim P(a\mid q)$. Without contextual grounding, the model's responses should exhibit higher variability due to unresolved ambiguities (e.g., unknown reference in the query).

To measure answer consistency, we cluster answers by their underlying meaning using bidirectional entailment \cite{su}. Specifically, two answers $a_1^{(c,q)}$ and $a_2^{(c,q)}$ are assigned to the same semantic cluster if and only if: (1) the contextualized meaning of $a_1^{(c,q)}$ logically entails $a_2^{(c,q)}$ and (2) conversely, $a_2^{(c,q)}$ also entails $a_1^{(c,q)}$. After applying clustering, the contextual answer set is partitioned into ${K}^{(c, q)}=\left\{K_{1}^{(c, q)}, K_{2}^{(c, q)}, \ldots, K_{\alpha}^{(c, q)}\right\}$. Similarly, the context-free answer set is partitioned into ${K}^{(q)}=\left\{K_{1}^{(q)}, K_{2}^{(q)}, \ldots, K_{\beta}^{(q)}\right\}$. Such a clustering reflects the dispersion of answers. We expect $A^{(c,q)}$ to form fewer clusters (high consensus) due to contextual guidance, while $A^{(q)}$ should yield more clusters (low consensus) reflecting uncertainty. 
Specifically, we use entropy to quantify answer consistency, where lower entropy indicates higher consensus. The key insight is that the entropy difference isolates context's contribution to consistency and arises solely from context availability since $q$ is identical in both conditions. Thus, $\Delta H$ quantifies how much context reduces uncertainty in answer generations.


\paragraph{Entropy Reduction Calculation}
We then compute the entropy difference between the two answer sets:
\begin{equation}
\begin{split}
    \Delta H  = & H({K}^{(q)})-H({K}^{(c, q)})\\ 
     = &\sum_{k\in K^{(c,q)}}^{} P(k|c,q)logP(k|c,q) - \\
     &\sum_{k\in K^{(q)}}^{} P(k|q)logP(k|q).
\end{split}
\end{equation}


The entropy reduction $\Delta H$ is based on Shannon's information theory, measuring how much the external context $c$ constrains the model's predictions for query $q$. Formally, the entropy difference can be rewritten as a conditional mutual information bound \cite{CMI}:

\begin{equation}
\Delta H = I(K^{(c,q)}; c | q) + \epsilon,
\end{equation}
where $I(K^{(c,q)}; c | q)$ represents the mutual information between context $c$ and the semantic clusters $K^{(c,q)}$ given $q$, and $\epsilon$ captures noise from sampling stochasticity. This formulation explicitly links $\Delta H$ to the contextual information gain, representing confidence. At the same time, we can quantify the data uncertainty through $-\Delta H$.

When $\Delta H \gg 0$ (i.e., $H({K}^{(q)}) \gg H({K}^{(c, q)})$), a strong positive $\Delta H$ implies that the context provides novel information that systematically reshapes the model's hypothesis space. As visualized in Figure~\ref{fig:method}, the context anchors the output distribution $P(a|c,q)$ to a low-entropy subspace that is distinct from the context-free distribution $P(a|q)$.

When $\Delta H \approx 0$, the entropy difference between $H({K}^{(q)})$ and $H({K}^{(c, q)})$ is negligible, implying that the conditioning on context $c$ does not meaningfully alter the uncertainty of the generated answers. This occurs in two distinct scenarios:
\begin{itemize}
    \item Both entropy values are low: The model’s intrinsic knowledge is sufficient to resolve query $q$ without relying on context $c$. For example, the factual question: \textit{``which coastline does Southern California touch?''} has a deterministic answer \textit{``Pacific''}, rendering context redundant. Here, the output space is already constrained, and model uncertainty (lack of knowledge) is minimal.
    \item Both entropy values are high: The knowledge implied by the model's own parameters is insufficient to correctly answer the question, and even if the context is provided, it is not effectively used or understood. In other words, it has high model uncertainty. For example, when asked \textit{``What was one of the Norman's major exports?''} with explicit context stating \textit{``normandy had been exporting fighting horsemen for more than a generation''}, the model generates high-variance outputs (\textit{``armor''}, \textit{``horses''}, \textit{``knights''}) rather than converging to \textit{``fighting horsemen''}. This reflects limited comprehension of contextual cues.


\end{itemize}

While $\Delta H$ identifies whether context reduces data uncertainty, it cannot alone distinguish between the two scenarios when $\Delta H \approx 0$. To address this, we introduce unified consistency as a complementary metric.

\subsubsection{Unified Consistency Measurement}
The second stage focuses on disambiguating cases with low 
$\Delta H$ by testing output stability under context perturbations. We assess the consistency of model outputs across context-conditioned answers ${A}^{(c,q)}$ and context-free answers ${A}^{(q)}$. If the unified answers ${A}^{global} = {A}^{(c,q)} \cup {A}^{(q)}$ exhibit low dispersion or high consensus, it indicates low model uncertainty, as outputs remain stable regardless of context variations. This confirms that the model has sufficient parametric knowledge to resolve the query independently. Conversely, high dispersion or low consensus indicates significant model uncertainty, revealing either: (1) inadequate parametric knowledge about the query domain, or (2) failure to effectively process and utilize contextual information. This diagnostic decomposition enables targeted improvements: cases showing low consensus highlight opportunities for model enhancement, while high-consensus results confirm the model's independent reasoning capability.

Following the previous work \cite{ecc}, we embed ${A}^{global}$ in a graph Laplacian where nodes represent answers and edge weights reflect pairwise semantic similarities. We can adopt either average pairwise distance or the average distance from the center as the unified consistency measure:

\begin{equation}
    G C_{\text {pairwise}}=\frac{1}{n(1-2n)} \sum_{1 \leq i<j \leq 2n} d\left(a^{global}_{i}, a^{global}_{j}\right),
\end{equation}
or
\begin{equation}
    G C_{\text {center}}=-\frac{1}{2n} \sum_{i=1}^{2n} d\left(A_{\text {center}}, a^{global}_{i}\right).
\end{equation}

\subsubsection{Neural Weighting Mechanism}
To dynamically fuse $\Delta H$ and $GC$ (either $G C_{\text {pairwise}}$ or $G C_{\text {center}}$) into a final confidence score, we train a 2-layer multi-layer perceptron (MLP) with ReLU activation:
\begin{equation}
\conf = \sigma\left(W_2 \cdot \text{ReLU}(W_1[\Delta H; GC] + b_1) + b_2\right).
\end{equation}

\section{Experiments}
\label{exp}
\begin{table*}[t]
	\centering
	\resizebox{0.9\textwidth}{!}{
		\begin{tblr}{
				cell{1}{3} = {c=5}{c},
				cell{1}{8} = {c=5}{c},
				cell{3-10}{1} = {c=1}{c},
				vline{3} = {1-9}{dotted},
				vline{8} = {1-9}{dotted},
				hline{1,10} = {-}{},
				hline{3} = {-}{dashed},
			}
			                      & & Llama-8B & & & & & Qwen-14B & & & & \\
			                      & & CoQA & SQuAD & QuAC & BioASQ & EduQG & CoQA & SQuAD & QuAC & BioASQ & EduQG \\
			Rouge\_L              & & 0.8476 & 0.8812 & 0.9074 & 0.8070 & 0.8756 & 0.7232 & 0.6841 & 0.7375 & 0.6104 & 0.8625 \\
			BLEU & & 0.8579 & 0.8776 & 0.9006 & 0.8353 & 0.8858 & 0.7608 & 0.7028 & 0.7444 & 0.6581 & 0.8479 \\
			Degree\_Matrix & & 0.8662 & 0.9135 & 0.9098 & 0.9013 & 0.9428 & 0.7074 & 0.7191 & 0.7398 & 0.6203 & 0.8789 \\
			Eccentricity & & 0.8671 & 0.8999 & 0.8966 & 0.8906 & 0.9369 & 0.7629 & 0.7435 & 0.7399 & 0.6726 & 0.8620 \\
			EigValLaplacian & & 0.8546 & 0.8926 & 0.9062 & 0.9005 & 0.8788 & 0.7480 & 0.7183 & 0.7354 & 0.6772 & 0.8088 \\
			NumSemSets & & 0.6761 & 0.6621 & 0.7449 & 0.6507 & 0.6383 & 0.5772 & 0.5412 & 0.5865 & 0.5887 & 0.5648 \\
			CRUX & & \textbf{0.8918} & \textbf{0.9166} & \textbf{0.9102} & \textbf{0.9364} & \textbf{0.9565} & \textbf{0.7845} & \textbf{0.7785} & \textbf{0.7530} & \textbf{0.7938} & \textbf{0.9055} \\
		\end{tblr}
	}
	\caption{AUROC score comparison between baselines and CRUX, with $n$ = 10.} 
	\label{tab:main_result}
\end{table*}

\begin{table*}[t]
	\centering
	\resizebox{0.9\textwidth}{!}{
		\begin{tblr}{
				cell{1}{3} = {c=5}{c},
				cell{1}{8} = {c=5}{c},
				cell{3}{1} = {r=2}{},
				cell{3}{3} = {l},
				cell{3}{4} = {l},
				cell{3}{5} = {l},
				cell{3}{6} = {l},
				cell{3}{7} = {l},
				cell{5}{1} = {r=2}{},
				vline{3} = {1-6}{dotted},
				vline{8} = {1-6}{dotted},
				vline{2} = {3-6}{dotted},
				hline{1,7} = {-}{},
				hline{3,5} = {-}{dashed},
			}
			&            & Llama-8B        &                 &        &        &        & Qwen-14B &         &        &        &             \\
			&            & CoQA            & SQuAD           & QuAC   & BioASQ & EduQG  & CoQA     & SQuAD   & QuAC   & BioASQ & EduQG       \\
			$\cruxwogc$     & w/o Clust. & 0.8532          & 0.8497          & 0.8489 & 0.7901 &0.7320 & 0.7764   & 0.7372  &0.7372  & 0.7487 & 0.8091      \\
			& w/ Clust.  & 0.8668          & 0.8914          & 0.8949 & 0.8907 & 0.7626 & 0.7543   & 0.7363  &0.7168  & 0.7559 & 0.8879      \\
			CRUX            & w/o Clust. & 0.8840          & 0.9028          & 0.8886 & 0.8966 & 0.9186 & \textbf{0.8025}   & \textbf{0.7922}  &0.7526  & 0.7580 & 0.8484      \\
			& w/ Clust.  & \textbf{0.8918} & \textbf{0.9166} & \textbf{0.9102} & \textbf{0.9364} & \textbf{0.9565} & 0.7845   & 0.7785  &\textbf{0.7530}  & \textbf{0.7938} & \textbf{0.9055}      
		\end{tblr}
	}
	\caption{AUROC score comparison for CRUX variants, with $n$ = 10. $\cruxwogc$ refers to CRUX without global consistency. ``w/ Clust.'' and ``w/o Clust.'' refer to whether clustering is or is not applied, respectively.}
	\label{tab:ablation}
\end{table*}

In this section we evaluate the quality of the confidence measures proposed in Section~\ref{crux}.

\subsection{Settings}
\paragraph{Datasets}
Building upon existing work \cite{su, ecc}, we use CoQA \cite{coqa}, an open-book question answering dataset where the model needs to leverage the given contextual evidence to answer questions. To enhance generalization, we integrate two widely adopted reading comprehension datasets, SQuAD \cite{squad} and QuAC \cite{quac} that share the paradigm of answering questions through context grounding. We filter questions in the datasets retaining all and only those that can be answered through explicit contextual information. 
To evaluate domain adaptation capabilities, we extend our investigation to specialized domains through two expert-curated datasets: BioASQ \cite{bioasq} (biomedical QA) and EduQG \cite{eduqg} (educational assessment QA). To ensure alignment with our experimental objectives, we retain yes/no and factoid questions for BioASQ and contexts of 1,000 to 2,000 words for EduQG.

\paragraph{Models} To evaluate the effectiveness and generalizability of our approach, we conduct experiments using two widely used language models: LLaMA-3-8B \cite{grattafiori2024llama3herdmodels} and Qwen-14B \cite{qwen2025qwen25technicalreport}, testing our method's robustness to model size variations. The selection of these open-source models ensures the reproducibility of our findings. 

\paragraph{Baseline Methods}
We compare six established methods spanning lexical and semantic dimensions for evaluating uncertainty.
 ROUGE and BLEU measure n-gram overlap consistency between generated and reference answers to quantify confidence. Degree Matrix, Eccentricity and Laplacian Eigenvalue use the graph Laplacian matrix to measure similarity dispersion, thus distinguishing confident answers.
NumSemSets counts distinct concept clusters in latent space to measure confidence beyond lexical matching.
\paragraph{Evaluation Metric} Following prior works \cite{su, ecc}, we formulate confidence estimation as a binary classification task: determining whether to trust a model-generated answer for a given question and context. We adopt the Area Under the Receiver Operating Characteristic curve (AUROC) as our primary evaluation metric. It measures the probability that a randomly chosen correct response receives higher confidence than an incorrect one, providing threshold-agnostic performance assessment.

\paragraph{Labeling} To evaluate the correctness of generated responses, we propose a robust inference-driven approach that takes advantage of natural language inference (NLI) and majority voting. For a given question, we assess each generation against the reference answer using an NLI model\footnote{We use DeBERTa-v3-base-mnli-fever-anli}. Specifically, we frame the reference answer as the premise and each generated response as the hypothesis, computing the probability that the response entails (correct) or contradicts (incorrect) the reference. Each generation is assigned a binary label (1 for entailment; 0 otherwise). The final correctness label is determined by max-vote aggregation. If the majority of generations are deemed correct, the collective output is labeled correct (1); otherwise, incorrect (0).

\begin{figure}[htbp]
	\centering
	\includegraphics[scale=0.23]{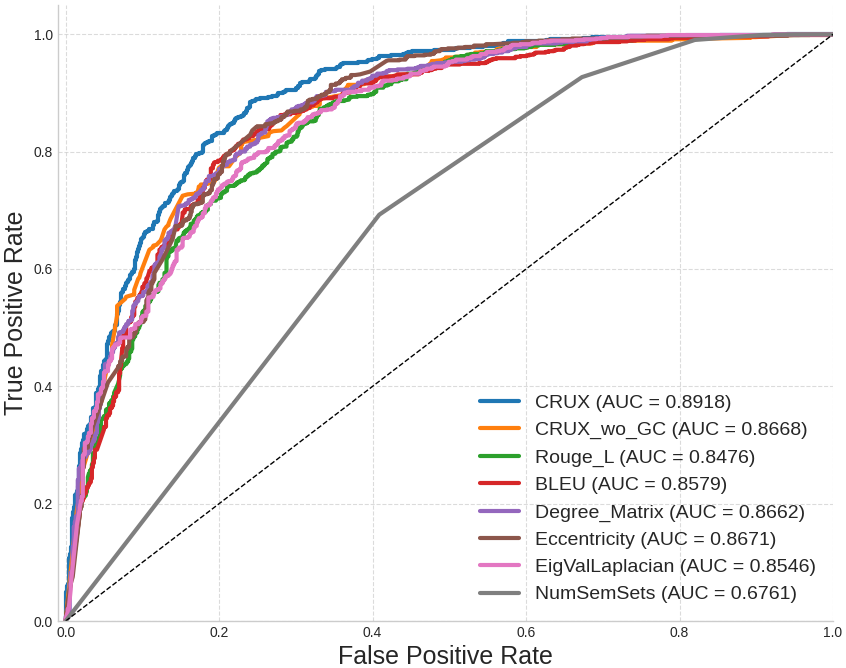}
	\caption{AUROC Curves for CoQA under Llama-8B}
	\label{fig:coqa_llama_auroc}
\end{figure}

\begin{figure}[htbp]
	\centering
	\includegraphics[scale=0.23]{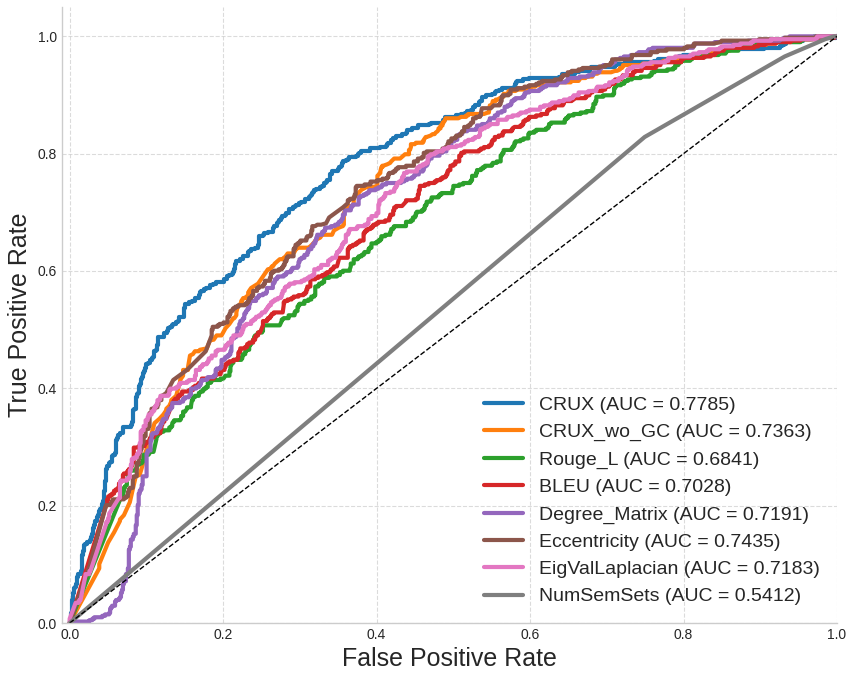}
	\caption{AUROC Curves for SQuAD under Qwen-14B}
	\label{fig:squad_qwen_auroc}
\end{figure}

\begin{figure*}[htbp]
	\centering
	\includegraphics[scale=0.40]{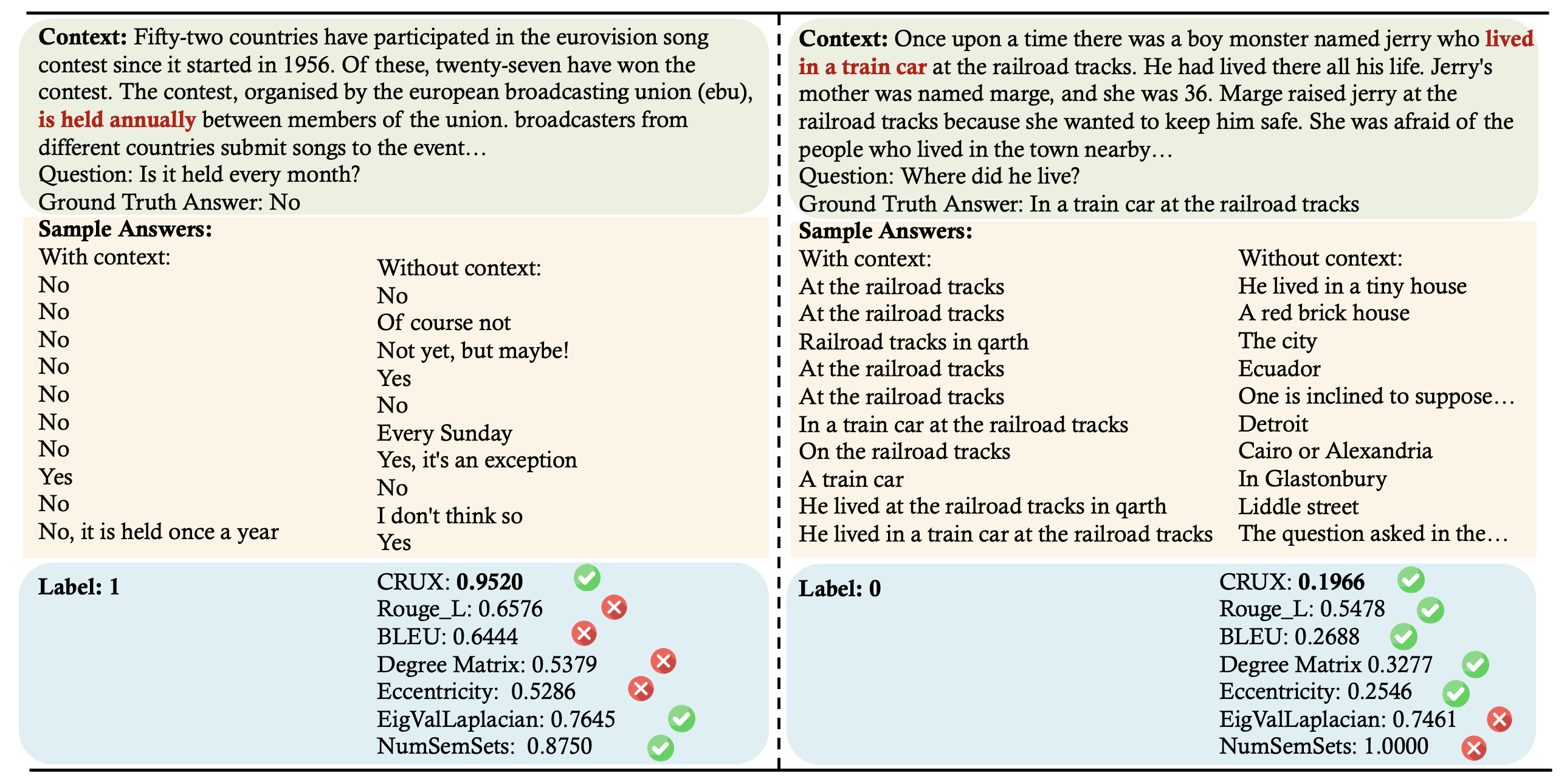}
	\caption{Case Study. The left panel (Case 1) demonstrates high-quality responses where context resolves confusion (label=1), while the right panel (Case 2) shows hallucination-prone answers where responses fail to answer the question (label=0).}
	\label{fig:cases}
\end{figure*}

\subsection{Results}
\paragraph{Overall Model Performance}
Table~\ref{tab:main_result} presents a comprehensive AUROC comparison between CRUX and the six baseline confidence estimation methods across five diverse QA datasets and two LLM architectures. The results reveal several key findings: First, CRUX consistently outperforms all baselines across both models and all datasets, achieving state-of-the-art AUROC scores (e.g., 0.9102 on QuAC with Llama). This demonstrates our method's superior ability to distinguish correct from incorrect predictions.

Second, Llama-8B consistently outperforms Qwen-14B across all metrics (e.g., Rouge\_L, BLEU, Degree\_Matrix) and datasets (e.g., CoQA, BioASQ). For instance, Llama achieves a CRUX's score of 0.9364 on BioASQ compared to Qwen’s 0.7938. This is notable given Llama’s smaller size (8B vs. 14B parameters). The performance gap between Llama-8B and Qwen-14B may primarily stem from differences in pre-training data and methodology. Llama-8B's advantage in English benchmarks likely arises from its focused training on publicly available online data in English, which aligns closely with the linguistic patterns in English QA benchmarks. In contrast, Qwen-14B's design prioritizes Chinese-language data and cross-lingual alignment, trading some English task specialization for broader multilingual coverage.

\paragraph{Ablation Study Insights}
To gain a deeper understanding of the contributions of CRUX's components, we conducted controlled ablation studies. Table~\ref{tab:ablation} provides critical insights into CRUX's components.

\textbf{Impact of Clustering in CRUX:} Clustering enhances Llama’s performance in all five datasets. For example, CRUX with clustering achieves Llama’s highest scores in BioASQ (0.9364 vs. 0.8966 without clustering). However, Qwen benefits less consistently from clustering, with improvements limited to specific datasets. In CoQA and SQuAD, clustering even degrades performance. It is mainly because that Qwen may generate outputs that appear similar but contain critical errors when lacking the necessary context, which leads to noisy results. A concrete example illustrates this: For the question in Figure~\ref{fig:method}, Qwen without context generates multiple incorrect responses like \textit{``Cotton is not a color, it is a natural fiber''}, \textit{``Cotton is not inherently a specific color''}, or \textit{``Cotton is not a color, but a natural fiber''}. Although semantically similar, these are distinct erroneous outputs. Crucially, clustering semantically similar incorrect answers together artificially reduces the entropy difference used to measure confidence, leading to an underestimation of reliability even with the gain of contextual information.

\textbf{Role of Unified Consistency:} The experimental results demonstrate that unified consistency significantly enhances the performance of both LLaMA-8B and Qwen-14B across diverse question-answering datasets. Crucially, this improvement builds upon the models' inherent, robust capabilities for question answering without relying on additional context. Both models fundamentally possess the ability to comprehend part of the questions and generate accurate responses. For an example in EduQG dataset: \textit{``What is a characteristic of financial accounting information?''} (with choices including \textit{``summarizes what has already occurred''}, \textit{``should be incomplete in order to confuse competitors''}, and \textit{``provides investors guarantees about the future''}). The models readily identify the correct answer (\textit{``summarizes what has already occurred''}) because this represents fundamental, widely-known accounting knowledge likely encountered extensively during pre-training. The incorrect choices contain obvious conceptual errors or implausible assertions, making them easily distinguishable by a model with a solid grasp of basic principles.

\paragraph{AUROC Curves Visualization} To further validate discriminative performance, we visualize 
AUROC curves for two representative settings. Figures~\ref{fig:coqa_llama_auroc} and \ref{fig:squad_qwen_auroc} compare CRUX with clustering (blue) against six baselines and an ablation study $\cruxwogc$ (orange) where global consistency is removed, under the two language models, respectively. For CoQA with Llama-8B, our approach achieves the best separability of correct/incorrect predictions ($\text{AUROC} = 0.8918$). Similarly, for SQuAD with Qwen-14B, our method ($\text{AUROC} = 0.7785$) outperforms all baselines 
Additionally, the figures show a significant performance drop in $\cruxwogc$ (orange), highlighting the critical importance of global consistency for discriminative ability.
\paragraph{Case Study}
Figure~\ref{fig:cases} illustrates two representative case studies that reveal differences between our approach and the baselines. 
In these cases, we set the confidence threshold to 0.7. We see that CRUX correctly gives the highest (lowest) confidence values for the correct (incorrect) answer, while the baselines incorrectly judge the answer in one of the cases.


In Case 1 (left), CRUX gives a confidence of 0.9520, which exceeds the 0.7 threshold, leading to a correct prediction. This is achieved by recognizing the significant entropy reduction from chaotic without-context responses (e.g., ``Every Sunday'', ``Yes, it's an exception'') to predominantly correct with-context answers. Moreover, it remains stable against both outliers (single ``Yes'') and acceptable elaborations (``No, it is held once a year''), proving robust to answer variations. Conversely, Rouge-L (0.6576), BLEU (0.6444), Degree Matrix (0.5379), and Eccentricity (0.5286) fall below the threshold due to oversensitivity to these minor variations. Thus, they fail to distinguish noise from valid answers. While EigValLaplacian and NumSemSets succeed when answers are uniformly correct, they fail when consistency disguises critical errors. Case 2 (right) exposes the fatal flaw of these methods, where EigValLaplacian (0.7461) and NumSemSets (1.0000) greatly exceed the threshold because they mistake surface-level agreement (``railroad tracks'') for true consistency. By incorporating global consistency measures, CRUX can alleviate this problem of consistency errors.

\section{Related Works}
\label{related work}

\subsection{Confidence/Uncertainty Estimation}
Traditional uncertainty quantification methods are usually based on Bayesian principles \cite{Bayesian_uq, heek2019bayesianinferencelargescale, KWON2020106816} by modeling output distributions or likelihoods. 
However, these approaches struggle in the realm of free-form text generation with LLMs, where token-level probabilities fail to reflect reliability \cite{ma2025estimatingllmuncertaintyevidence} and commercial LLMs are closed-source, precluding access to internal probabilities \cite{yona-etal-2024-large}. To address these challenges, recent works focus on LLMs consistency-based methods \cite{su, ecc, luq}, which measure agreement across multiple generations, and self-evaluation methods \cite{lin2022teachingmodelsexpressuncertainty, xiong2024llmsexpressuncertaintyempirical, heo2025llmsestimateuncertaintyinstructionfollowing}, where LLMs assess their own confidence. 
However, both paradigms neglect contextual faithfulness, which is the degree to which outputs are derived from provided context rather than from memorized knowledge. This gap is particularly problematic in context-dependent scenarios, especially within specialized domains such as legal applications \cite{legal_qbr}. To address this gap, our work considers the context information gain to measure contextual faithfulness. In addition, we explicitly disentangle epistemic uncertainty from aleatoric uncertainty, which advances beyond consistency-centric singularity and opacity of self-assessment, providing a grounded solution for context-aware confidence estimation.

\subsection{Contrastive Decoding Methods}
Traditional decoding methods for text generation, such as greedy search and sampling \cite{info12090355}, often prioritize likelihood but struggle to balance fluency, coherence, and contextual faithfulness. Recent contrastive decoding methods address these issues by leveraging differences between model behaviors.
\cite{li-etal-2023-contrastive} proposed Contrastive Decoding (CD), contrasting outputs from large expert and small amateur language models to suppress repetitive or incoherent text. 
Extensions to reasoning tasks \cite{obrien2023contrastivedecodingimprovesreasoning} improved performance on benchmarks like GSM8K by reducing reasoning errors.
To enhance context dependence, methods like Context-Aware Decoding (CAD) \cite{shi-etal-2024-trusting} contrast outputs with and without context, downweighting tokens conflicting with external evidence. 
In addition, Decoding with Generative Feedback (DeGF) \cite{zhang2025selfcorrectingdecodinggenerativefeedback} mitigates hallucinations in vision-language models by contrasting token predictions conditioned on original and synthesized images.
Inspired by those work, we adopt a contrastive decoding method to measure model epistemic uncertainty in CQA tasks.

\section{Limitation}

The framework relies on an LLM's ability of effectively utilizing context $c$ to refine its output space. However, weaker models may fail to extract or integrate contextual signals. For instance, if a model lacks basic capabilities, even relevant context may not reduce $H({K}^{(c, q)})$, skewing $\Delta H$ interpretations. In fact, other uncertainty estimation methods (such as those leveraging self-evaluation) also require high-performing LLMs. Weaker models risk generating hallucinations or repeating consistency errors.

\section{Conclusion}

In this work, we propose CRUX, a dual-metric framework that quantifies confidence through contextual information gain and global consistency, unified by a neural network-based dynamic weighting mechanism. Experiments demonstrate that CRUX significantly outperforms existing methods across diverse datasets, including domain-specific scenarios such as biomedical and education. 

While CRUX provides a robust foundation for context-aware confidence estimation, several promising directions remain: (1) Integration with Retrieval-Augmented Generation (RAG): Current method assumes context is pre-provided and informative, but real-world CQA often requires dynamic context retrieval. By incorporating RAG, we could jointly evaluate confidence in both the retrieved evidence (e.g., document relevance, source reliability) and the generated answers, necessitating adaptive weighting between retrieval and generation modules.
(2) As LLMs are increasingly capable of handling long-form contexts, CRUX could decompose confidence at the claim level to enhance interpretability. For example, in a generated answer that contains multiple factual claims (e.g., ``He was born in 1911 [Claim 1], and he loves art. [Claim 2]''), contextual information gain and global consistency could be computed per claim. This would enable error localization (e.g., Claim 2 has high aleatoric uncertainty) and allow users to trace confidence back to specific context segments.

\newpage
\bibliography{aaai2026}
\newpage

\end{document}